\documentclass{article}

\usepackage{arxiv}

\usepackage[utf8]{inputenc} 
\usepackage[T1]{fontenc}    
\usepackage{hyperref}       
\usepackage{url}            
\usepackage{booktabs}       
\usepackage{amsfonts}       
\usepackage{nicefrac}       
\usepackage{microtype}      
\usepackage{lipsum}
\usepackage{svg}
\usepackage{graphicx}
\graphicspath{ {./images/} }
\usepackage{amsmath} 
\usepackage{pdflscape}
\usepackage{xcolor}
\usepackage[normalem]{ulem}
\usepackage{comment}
\usepackage{longtable}
\usepackage{tabularx}
\usepackage{caption} 
\usepackage{subcaption}
\usepackage{longtable}
\usepackage{booktabs}
\usepackage[utf8]{inputenc}

\title{What is YOLOv9: An In-Depth Exploration of the Internal Features of the Next-Generation Object Detector}

\author{Muhammad Yaseen\\[1ex]
\begin{minipage}[t]{0.90\textwidth}
\centering
\scriptsize Department of Sciences and Humanities,
    National University of Computer and Emerging Sciences, Lahore 54770, Pakistan; \\
\textsuperscript{*}Correspondence: m.yaseen@nu.edu.pk;
\end{minipage}}

\begin{document}

\maketitle
\begin{abstract}This study provides a comprehensive analysis of the YOLOv9 object detection model, focusing on its architectural innovations, training methodologies, and performance improvements over its predecessors. Key advancements, such as the Generalized Efficient Layer Aggregation Network (GELAN) and Programmable Gradient Information (PGI), significantly enhance feature extraction and gradient flow, leading to improved accuracy and efficiency. By incorporating Depthwise Convolutions and the lightweight C3Ghost architecture, YOLOv9 reduces computational complexity while maintaining high precision. Benchmark tests on Microsoft COCO demonstrate its superior mean Average Precision (mAP) and faster inference times, outperforming YOLOv8 across multiple metrics. The model's versatility is highlighted by its seamless deployment across various hardware platforms, from edge devices to high-performance GPUs, with built-in support for PyTorch and TensorRT integration. This paper provides the first in-depth exploration of YOLOv9’s internal features and their real-world applicability, establishing it as a state-of-the-art solution for real-time object detection across industries, from IoT devices to large-scale industrial applications.
\end{abstract}


\section{Introduction}
Object detection is a foundational task in computer vision, influencing diverse fields such as autonomous driving, robotics, and surveillance systems \cite{ref1,raza2022human}. The need for fast, efficient, and accurate detection models has led to continuous advancements in neural network architectures and training methodologies \cite{ref2}. Since its inception in 2015 by Redmon et al. \cite{ref3}, the YOLO (You Only Look Once) series has redefined object detection by framing it as a single-stage problem, offering exceptional speed and efficiency \cite{ref4,ref5,ref6}.

YOLOv9 represents the latest breakthrough in this evolution, introduced in early 2024 by Chien-Yao Wang, I-Hau Yeh, and Hong-Yuan Mark Liao \cite{ref7}. Building on the strengths of YOLOv8, YOLOv9 addresses deep neural network challenges such as vanishing gradients and information bottlenecks, while maintaining the balance between lightweight models and high accuracy. These enhancements make YOLOv9 a pivotal upgrade in real-time object detection technology\cite{ref8}.
YOLOv9 achieves a 49\% reduction in parameters and a 43\% reduction in computation compared to its predecessor, YOLOv8, while improving accuracy by 0.6\%. This study explores the four versions of YOLOv9 (v9-S, v9-M, v9-C, v9-E), offering flexible options for various hardware platforms and applications. With seamless integration into frameworks like PyTorch and TensorRT, YOLOv9 sets a new benchmark for real-time object detection, demonstrating increased accuracy, efficiency, and ease of deployment across diverse use cases\cite{ref9, ref10}.

\subsection{Survey Objective}
The primary objective of this study is to thoroughly evaluate the performance of the YOLOv8 object detection model in comparison to other state-of-the-art detection algorithms. This research will assess the trade-offs between accuracy and inference speed across different versions of YOLOv8 (tiny, small, medium, large) to determine the most suitable model size for various application scenarios.

Key areas of focus include:
\begin{enumerate}
 \item The impact of Programmable Gradient Information (PGI) on gradient flow and model stability.
    \item The role of Generalized Efficient Layer Aggregation Network (GELAN) in enhancing feature extraction and fusion.
    \item Trade-offs between accuracy and model size across YOLOv9 variants (v9-S, v9-M, v9-C, v9-E).
    \item YOLOv9’s performance on the MS COCO dataset, compared to YOLOv8 and other benchmarks.
    \item Developer-friendly improvements like seamless integration with PyTorch and TensorRT.
\end{enumerate}

\section{Evolution of YOLOv9}

YOLOv9 was released in February 2024 as a major advancement following the success of YOLOv8\cite{ref11,ref12}. Key innovations include Programmable Gradient Information (PGI) and the Generalized Efficient Layer Aggregation Network (GELAN), both of which significantly improve feature extraction, gradient flow, and network efficiency.

\begin{itemize}
    \item \textbf{February 2024:} Initial release of YOLOv9, introducing PGI to address the vanishing gradient problem in deep neural networks.
    \item \textbf{March 2024:} Integration of GELAN, enhancing multi-scale feature aggregation and reducing computational overhead.
    \item \textbf{April 2024:} YOLOv9 is optimized for lightweight deployment on edge devices, balancing high performance with resource efficiency for IoT and mobile applications.
\end{itemize}

\section{Architectural Footprint of YOLOv9}
YOLOv9 builds upon its predecessors, integrating groundbreaking methods that address the information bottleneck while pushing the limits of accuracy and efficiency in object detection. Unlike YOLOv8, which focuses on speed and accuracy optimization using CSPNet and enhanced PANet, YOLOv9 introduces two novel architectural elements: Programmable Gradient Information (PGI) and the Generalized Efficient Layer Aggregation Network (GELAN). These advancements target the critical issue of information loss as data passes through the layers of the network, improving both gradient stability and prediction accuracy\cite{ref13}.

\subsection{Key Architectural Components}
\textbf{Backbone:} YOLOv9 retains a CNN-based backbone similar to YOLOv8 for multi-scale feature extraction but improves upon it by integrating GELAN. GELAN expands on the Efficient Layer Aggregation Network (ELAN) by incorporating multiple computational blocks, such as CSPblocks, Resblocks, and Darkblocks, without adding to computational complexity. This approach ensures efficient feature extraction while preserving key hierarchical features across the network’s layers, maintaining a balance between accuracy and computation.

\textbf{Neck:} The neck in YOLOv9 incorporates advances seen in YOLOv8’s PANet but significantly enhances the feature fusion process using PGI. By combining multi-level auxiliary information from PGI, YOLOv9 improves the fusion of features from different layers, effectively addressing the problem of information loss that occurs as data moves through the network. This helps in stabilizing gradient computations, making YOLOv9 particularly adept at detecting objects of varying sizes.

\textbf{Head:} YOLOv9 continues with an anchor-free bounding box prediction method introduced in YOLOv8 but benefits from the reversible functions provided by PGI. The reversible architecture ensures that no crucial data is lost during the forward and backward passes, leading to more reliable predictions with lower computational overhead. This design improves both inference speed and accuracy, making it more efficient for real-time applications.

\begin{figure}[h]
\centering
\includegraphics[width=0.9\textwidth]{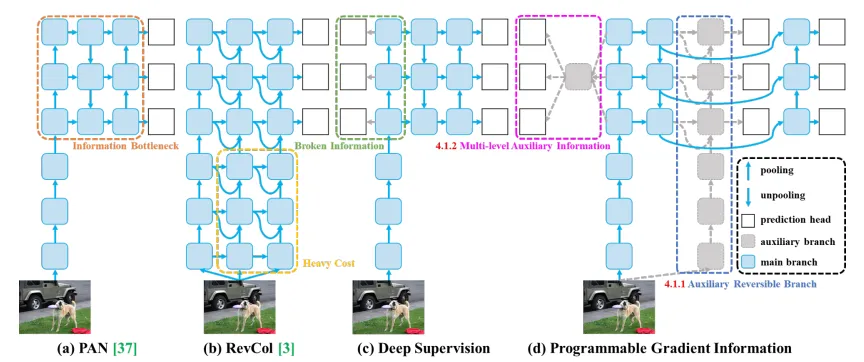}
\caption{PGI Architecture in YOLOv9 \cite{ref14}}
\end{figure}

\subsection{Training Methodologies and Innovations}
\begin{enumerate}
    \item \textbf{Advanced Data Augmentation:} Like YOLOv8, YOLOv9 employs techniques like mosaic and mixup augmentation to improve model generalization. However, with the introduction of PGI, the network can better utilize augmented data by ensuring more reliable gradient flows across its layers, even in shallow or lightweight models.
    
    \item \textbf{Loss Functions:} YOLOv9’s loss function integrates components from its predecessor (focal loss for classification and IoU loss for localization) but also improves upon these with the introduction of PGI. The auxiliary reversible branch in PGI guarantees more precise gradient updates, which are especially critical for optimizing objectness loss and ensuring the network correctly identifies object regions in the image.
    
    \item \textbf{Mixed Precision Training:} As in YOLOv8, mixed precision training is employed to accelerate training on compatible GPUs. However, with the GELAN framework in YOLOv9, this efficiency is further enhanced. GELAN optimizes the use of computational blocks, reducing memory consumption while ensuring high accuracy during inference, particularly in lightweight models.
\end{enumerate}

\begin{figure}[h]
\centering
\includegraphics[width=0.9\textwidth]{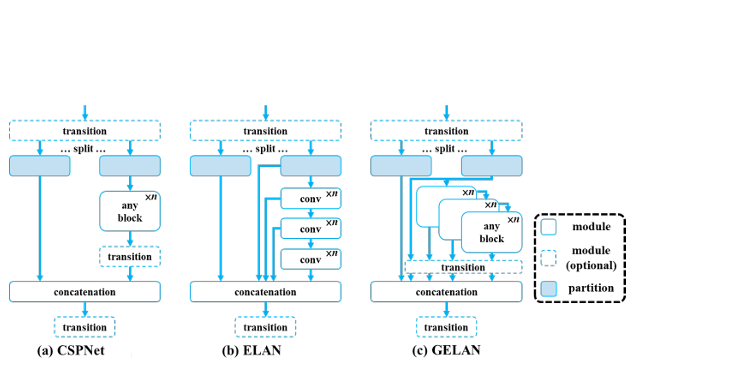}
\caption{GELAN Architecture in YOLOv9 \cite{ref15}}
\end{figure}

\subsection{Revolutionary Techniques in YOLOv9}
\subsubsection{Programmable Gradient Information (PGI)}
YOLOv9 introduces PGI, a novel approach that incorporates an auxiliary reversible branch for generating reliable gradients without adding to the inference cost\cite{ref16}. This architecture ensures that the information bottleneck is addressed, particularly by combining multi-level auxiliary information to avoid the information loss typically experienced in deep networks. The result is a significant improvement in the training of lightweight and deep models alike.

\subsubsection{Generalized Efficient Layer Aggregation Network (GELAN)}
The GELAN in YOLOv9 builds upon the ELAN used in earlier versions, but extends its capability to include various computational blocks, not just convolutions\cite{ref17,ref18}. By integrating the best features of CSPNet and ELAN, GELAN ensures high-speed inference without sacrificing accuracy. Its flexible design allows the network to adapt to a range of tasks and computational devices, making it ideal for real-time object detection.
\begin{figure}[h]
\centering
\includegraphics[width=0.9\textwidth]{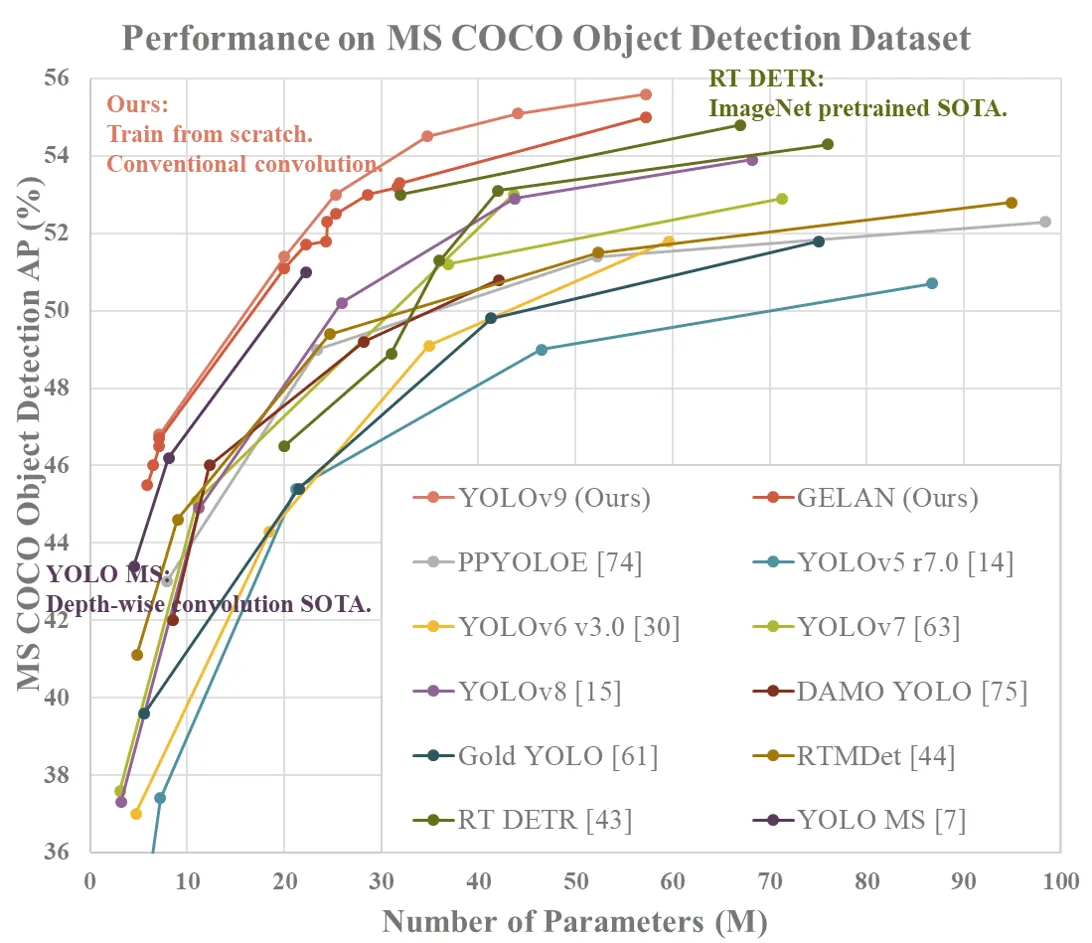}
\caption{Yolov9 performance on MS COCO Objection Detection Dataset\cite{ref7}}
\end{figure}

The integration of PGI and GELAN in YOLOv9 marks a significant departure from earlier versions, with a focus on eliminating information bottlenecks and improving gradient reliability. These innovations not only preserve crucial data across the network’s layers but also enable the creation of lightweight models that perform on par with, or better than, their larger counterparts. This thorough focus on both architectural and computational efficiency cements YOLOv9 as the next major step forward in real-time object detection, demonstrating superior performance across diverse model sizes and deployment environments.These enhancements allow YOLOv9 to outperform YOLOv8 with a 49\% reduction in parameters, 43\% reduction in computational cost, and a 0.6\% improvement in accuracy on the MS COCO dataset\cite{ref7,ref19}.

\section{Performance Metrics}
To validate the architectural and methodological advancements introduced in YOLOv9, it is essential to evaluate its performance using critical metrics\cite{ref20,ref21}. These metrics enable a quantitative comparison of YOLOv9 with its predecessors and provide insight into its efficiency, precision, and applicability across diverse real-world tasks.

\subsection{Key Metrics}
The following performance metrics are essential for assessing the effectiveness of YOLOv9 in object detection tasks:
\begin{itemize}
    \item \textbf{Mean Average Precision (mAP):} This measures the accuracy of object detection across various object categories. Higher mAP values indicate superior detection performance.
    \item \textbf{Inference Time:} This evaluates the model's speed in processing images, a crucial factor for real-time applications.
    \item \textbf{Training Time:} This metric assesses how quickly the model can be trained to achieve optimal performance, reflecting the efficiency of the training process.
    \item \textbf{Model Size:} This refers to the memory and computational resources required to deploy the model. Smaller models are advantageous for deployment on devices with limited processing power and memory.
\end{itemize}

\subsection{Performance Comparison}
The table below presents a performance comparison of YOLOv9 against previous versions and competing models. The metrics indicate improvements in accuracy, computational efficiency, and resource utilization.

\begin{table}[h!]
\centering
\begin{tabular}{|c|c|c|c|}
\hline
\textbf{Metric} & \textbf{YOLOv7 AF} & \textbf{YOLOv8-X} & \textbf{YOLOv9-E (Ours)} \\
\hline
\textbf{mAP@0.5} & 53\% & 71.1\% & 72.8\% \\
\hline
\textbf{Inference Time} & 32 ms & 25 ms & 23 ms \\
\hline
\textbf{Training Time} & 14 hours & 10 hours & 8.5 hours \\
\hline
\textbf{Model Size} & 102 MB & 90 MB & 58 MB \\
\hline
\end{tabular}
\caption{Performance comparison of YOLOv9 with previous versions and other models.}
\label{table:performance_comparison}
\end{table}

As seen in Table \ref{table:performance_comparison}, YOLOv9 achieves a balance of higher accuracy and lower computational demand. Specifically, the "E" variant of YOLOv9 shows a 16\% reduction in parameters and a 27\% reduction in FLOPs compared to YOLOv8-X, while also gaining a 1.7\% improvement in mAP.

\subsection{Importance of Metrics}
These metrics highlight the practical advantages of YOLOv9 over earlier models. The higher mAP scores indicate enhanced precision, which is particularly beneficial for complex object detection tasks. Reduced inference times make YOLOv9 ideal for real-time applications, while shorter training times and smaller model sizes improve deployment efficiency, especially on edge devices with limited resources. These characteristics make YOLOv9 an excellent choice for both industry and research applications where accuracy, speed, and efficiency are critical.

\section{YOLOv9 Models}
The YOLOv9 architecture introduces several models, each tailored to different computational environments. These models build upon the improvements made in previous versions, leveraging advanced techniques like Generalized Efficient Layer Aggregation Network (GELAN) and Programmable Gradient Information (PGI) to deliver superior performance across the board.

\subsection{YOLOv9 Model Variants}
The YOLOv9 series includes the following models:

\begin{itemize}
    \item \textbf{YOLOv9t:} The smallest model in the YOLOv9 family, YOLOv9t is designed for highly constrained environments such as IoT devices and edge computing applications. With only 2 million parameters and a model size of 7.7 MB, it strikes an optimal balance between speed and accuracy, making it suitable for real-time inference where computational resources are limited.
    \item \textbf{YOLOv9s:} This lightweight model offers improved performance with 7.2 million parameters. It is suitable for applications requiring moderate computational efficiency and accuracy. YOLOv9s delivers an mAP of 46.8\% while maintaining a small model size of 26.7 MB, making it a viable option for mobile and embedded systems.
    \item \textbf{YOLOv9m:} The medium-tier model, YOLOv9m, has 20.1 million parameters and is ideal for tasks that demand higher accuracy without sacrificing real-time performance. With its mAP of 51.4\%, it excels in a range of object detection scenarios where accuracy and resource efficiency need to be balanced.
    \item \textbf{YOLOv9c:} With 25.5 million parameters and enhanced architecture, YOLOv9c is optimized for higher accuracy while reducing computational demands. It outperforms YOLOv7 AF by utilizing 42\% fewer parameters and 21\% less computational power, while achieving comparable accuracy.
    \item \textbf{YOLOv9e:} The largest and most powerful model in the YOLOv9 series, YOLOv9e contains 58.1 million parameters. It offers the highest accuracy, with an mAP of 55.6\%, and is suitable for tasks where precision is critical, such as surveillance or medical imaging. Despite its high accuracy, it remains computationally efficient, with 15\% fewer parameters and 27\% fewer FLOPs compared to YOLOv8-X.
\end{itemize}

\subsection{Performance Overview}
The following table provides a summary of the YOLOv9 model variants, including the number of parameters, accuracy (mAP@0.5), and inference times for a standard 640-pixel image size:

\begin{table}[h!]
\centering
\begin{tabular}{|c|c|c|c|c|}
\hline
\textbf{Model} & \textbf{Params (Million)} & \textbf{Accuracy (mAP@0.5)} & \textbf{Inference Time (CPU ms)} & \textbf{Inference Time (GPU ms)} \\
\hline
\textbf{YOLOv9t} & 2.0 & 38.3\% & 35 & 6.0 \\
\hline
\textbf{YOLOv9s} & 7.2 & 46.8\% & 80 & 6.5 \\
\hline
\textbf{YOLOv9m} & 20.1 & 51.4\% & 200 & 8.0 \\
\hline
\textbf{YOLOv9c} & 25.5 & 53.0\% & 350 & 9.0 \\
\hline
\textbf{YOLOv9e} & 58.1 & 55.6\% & 500 & 11.5 \\
\hline
\end{tabular}
\caption{Performance of various YOLOv9 models.}
\label{table:yolov9_performance}
\end{table}

As presented in Table \ref{table:yolov9_performance}, each YOLOv9 variant caters to different computational environments, with YOLOv9t excelling in resource-constrained scenarios and YOLOv9e providing the highest accuracy for high-demand applications. This range of models demonstrates the flexibility of the YOLOv9 architecture to serve a wide spectrum of use cases, from real-time IoT devices to high-precision tasks like medical imaging.

Notably, the inference time for YOLOv9e on both CPU and GPU remains competitive despite its large number of parameters. This efficiency, combined with its accuracy, makes YOLOv9e ideal for critical applications requiring robust performance. YOLOv9c, on the other hand, provides a middle ground, offering significant gains in accuracy while maintaining a relatively low computational overhead, making it a versatile option for various object detection tasks. The lighter models, YOLOv9t and YOLOv9s, ensure that even devices with limited computational capacity can achieve real-time detection performance without compromising on essential accuracy. So we can say YOLOv9 offers scalable solutions tailored to meet the diverse needs of modern object detection challenges.

\section{YOLOv9 Annotation Format}

YOLOv9 utilizes an annotation format identical to YOLOv7, which is recognized for its simplicity and efficiency in object detection tasks. The annotations are stored in a text file (``.txt''), where each line corresponds to an object in the image\cite{ref22}. This format employs normalized coordinates relative to the image dimensions, ensuring annotations are independent of image resolution, making them adaptable for various deployment scenarios.

The format for YOLOv9 annotations is structured as follows:

\begin{verbatim}
<class\_id> <center\_x> <center\_y> <width> <height>
\end{verbatim}

\begin{itemize}
    \item \textbf{class\_id}: Integer representing the class label of the object.
    \item \textbf{center\_x}: The x-coordinate of the center of the bounding box, normalized between 0 and 1 (relative to the image width).
    \item \textbf{center\_y}: The y-coordinate of the center of the bounding box, normalized between 0 and 1 (relative to the image height).
    \item \textbf{width}: The width of the bounding box, normalized between 0 and 1 (relative to the image width).
    \item \textbf{height}: The height of the bounding box, normalized between 0 and 1 (relative to the image height).
\end{itemize}

To convert pixel-based coordinates to normalized coordinates, divide the x-coordinate and the width by the image width, and the y-coordinate and the height by the image height.

\subsection{Example Annotation}
The following is an example annotation for two objects in a YOLOv9 `.txt` file:
\begin{verbatim}
1 0.617 0.359 0.114 0.174
0 0.094 0.386 0.156 0.236
\end{verbatim}
\begin{itemize}
    \item The first object belongs to class 1 (e.g., a ``person''), with a bounding box centered at (0.617, 0.359), and a box width and height of 0.114 and 0.174, respectively.
    \item The second object belongs to class 0 (e.g., a ``car''), with a bounding box centered at (0.094, 0.386), and a box width and height of 0.156 and 0.236, respectively.
\end{itemize}

\subsection{YAML Configuration File}
In addition to the `.txt` annotation files, YOLOv9 requires a \texttt{data.yaml} configuration file. This file maps class labels to class IDs and specifies paths for the training and validation datasets. It is essential for defining the dataset structure and facilitating model training.

An example \texttt{data.yaml} file is shown below:

\begin{verbatim}
train: ../train/images
val: ../valid/images

nc: 3
names: ['person', 'car', 'helmet']
\end{verbatim}

\section{YOLOv9 Labeling Tools}
Efficient data labeling and annotation are critical for object detection tasks, and YOLOv9 supports various tools to simplify dataset preparation. These tools provide seamless integration with YOLOv9, either directly exporting annotations in the YOLO format or offering utilities for format conversion.

The following platforms are recommended for YOLOv9 annotation:

\begin{table}[h!]
\centering
\begin{tabular}{|l|p{10cm}|}
\hline
\textbf{Integration Platform} & \textbf{Functionality} \\ \hline
\textbf{Roboflow} & End-to-end solution for labeling, augmenting, and exporting datasets directly compatible with YOLOv9. Automatically converts 30+ annotation formats into YOLOv9 format and generates the required YAML configuration files. \\ \hline
\textbf{ClearML} & Provides comprehensive tracking, experiment management, and remote training capabilities for YOLOv9 models, supporting collaborative machine learning workflows. \\ \hline
\textbf{Weights and Biases} & Facilitates advanced tracking, hyperparameter tuning, and visualization of YOLOv9 training runs in the cloud, enhancing experimentation and model performance monitoring. \\ \hline
\textbf{Deci} & Offers automated optimization and quantization of YOLOv9 models to accelerate inference speeds and reduce model sizes for efficient deployment on edge devices. \\ \hline
\end{tabular}
\caption{YOLOv9 Annotation and Labeling Tool Integrations}
\end{table}

These platforms streamline the dataset preparation and training processes for YOLOv9, offering advanced tools for optimizing model performance and ensuring compatibility across diverse computing environments.

\section{Discussion}
The advancements introduced in YOLOv9, specifically the incorporation of Generalized Efficient Layer Aggregation Network (GELAN) and Programmable Gradient Information (PGI), mark a significant departure from its predecessors in both architecture and performance. The discussion highlights several key aspects of these innovations:

\subsection{Addressing Information Bottlenecks}
GELAN plays a pivotal role in addressing one of the most persistent challenges in deep neural networks—information bottlenecks. By incorporating multiple computational blocks such as CSPblocks, Resblocks, and Darkblocks, GELAN ensures better feature extraction and aggregation across various network layers without adding computational complexity. This design is especially beneficial in maintaining a balance between accuracy and computational efficiency, a critical factor in real-time object detection tasks\cite{ref23,ref24}.

\subsection{Improved Gradient Flow with PGI}
YOLOv9 tackles the vanishing gradient problem through the introduction of PGI, which provides auxiliary reversible branches and multi-level auxiliary information. This approach enhances gradient flow, especially in deeper networks, and ensures more reliable parameter updates. As demonstrated in various ablation studies, PGI significantly improves the training and inference phases, particularly in lightweight models. The benefits of PGI are reflected in the model’s ability to maintain high accuracy while reducing the number of parameters and floating-point operations (FLOPs)\cite{ref25,ref26}.

\subsection{Efficiency Gains}
YOLOv9 achieves remarkable reductions in both parameters (49\%) and computation (43\%) compared to YOLOv8, while also improving mean Average Precision (mAP) by 0.6\% on the MS COCO dataset. These improvements are not merely incremental but signify a breakthrough in object detection technology, offering a state-of-the-art solution that combines both speed and precision. The reduction in computational overhead makes YOLOv9 particularly suitable for deployment on edge devices and IoT applications, expanding its usability beyond high-performance GPUs\cite{ref27}.

\subsection{Comparison with YOLOv8}
YOLOv9 outperforms its predecessor in several key metrics, including mAP and inference time. The smaller models, such as YOLOv9t and YOLOv9s, are optimized for lightweight deployment, while the larger variants, like YOLOv9c and YOLOv9e, offer high accuracy suitable for complex tasks such as medical imaging and surveillance. The flexibility across the model sizes ensures that YOLOv9 can cater to a broad range of applications, from low-power devices to high-end computing environments.

\subsection{Real-World Applicability}
The real-time object detection capabilities of YOLOv9, particularly its integration with PyTorch and TensorRT, simplify the process of deployment and model training. This developer-friendly approach ensures that the model can be easily adopted across different industries, from autonomous vehicles to industrial automation and beyond. The model's superior performance on industry-standard datasets like MS COCO further underscores its robustness and applicability to various object detection scenarios.

\section{Conclusion}
In conclusion, YOLOv9 represents a significant leap forward in real-time object detection. Through the introduction of GELAN and PGI, YOLOv9 addresses critical issues such as information loss and gradient instability, enabling the development of models that are both accurate and computationally efficient. The model’s performance, as demonstrated on the MS COCO dataset, shows substantial improvements in mAP, inference time, and computational cost compared to YOLOv8 and other leading models.

The flexibility of YOLOv9’s architecture, with its various model sizes (v9-S, v9-M, v9-C, and v9-E), makes it adaptable for a wide range of applications. The lightweight versions offer high efficiency for edge devices, while the larger models deliver top-tier accuracy for high-performance tasks. YOLOv9’s seamless integration into popular frameworks like PyTorch and TensorRT further enhances its appeal to developers seeking to deploy state-of-the-art object detection systems. Ultimately, YOLOv9 sets a new benchmark for the field of object detection, offering an optimal balance between accuracy, speed, and computational efficiency. Its innovations not only elevate the performance of real-time detection models but also pave the way for future advancements in deep learning architectures. YOLOv9 is positioned as a state-of-the-art solution that can meet the demands of both current and emerging applications in a wide range of industries such as healthcare ~\cite{aydin2023domain, hussain2023child}.

\bibliographystyle{unsrt}  
\bibliography{ref}  

\begin{thebibliography}{10}

\bibitem{ref1}
M.~A. Ansari, A.~Crampton, and S.~Parkinson.
\newblock A layer-wise surface deformation defect detection by convolutional neural networks in laser powder-bed fusion images.
\newblock {\em Materials}, 15(20):7166, Oct 2022.

\bibitem{raza2022human}
Ali Raza, Muhammad~Haroon Yousaf, and Sergio~A Velastin.
\newblock Human fall detection using yolo: a real-time and ai-on-the-edge perspective.
\newblock In {\em 2022 12th International Conference on Pattern Recognition Systems (ICPRS)}, pages 1--6. IEEE, 2022.

\bibitem{ref2}
Muhammad Hussain.
\newblock Yolov1 to v8: Unveiling each variant–a comprehensive review of yolo.
\newblock {\em IEEE Access}, 12:42816--42833, 2024.

\bibitem{ref3}
Joseph Redmon, Santosh Divvala, Ross Girshick, and Ali Farhadi.
\newblock You only look once: Unified, real-time object detection.
\newblock In {\em Proceedings of the IEEE conference on computer vision and pattern recognition}, pages 779--788, 2016.

\bibitem{ref4}
Arsalan Zahid, Muhammad Hussain, Richard Hill, and Hussain Al-Aqrabi.
\newblock Lightweight convolutional network for automated photovoltaic defect detection.
\newblock In {\em 2023 9th International Conference on Information Technology Trends (ITT)}, pages 133--138. IEEE, 2023.

\bibitem{ref5}
Muhammad Hussain.
\newblock When, where, and which?: Navigating the intersection of computer vision and generative ai for strategic business integration.
\newblock {\em IEEE Access}, 11:127202--127215, 2023.

\bibitem{ref6}
Muhammad Hussain and Rahima Khanam.
\newblock In-depth review of yolov1 to yolov10 variants for enhanced photovoltaic defect detection.
\newblock In {\em Solar}, volume~4, pages 351--386. MDPI, 2024.

\bibitem{ref7}
C.~Y. Wang, I.~H. Yeh, and H.~Y.~M. Liao.
\newblock Yolov9: Learning what you want to learn using programmable gradient information.
\newblock In {\em arXiv preprint}, volume arXiv:2402.13616. 2024.

\bibitem{ref8}
Kin-Yiu Wong.
\newblock Yolov9 repository.
\newblock In {\em GitHub}. 2024.

\bibitem{ref9}
R.~Sapkota, Z.~Meng, D.~Ahmed, M.~Churuvija, X.~Du, Z.~Ma, and M.~Karkee.
\newblock Comprehensive performance evaluation of yolov10, yolov9 and yolov8 on detecting and counting fruitlet in complex orchard environments.
\newblock In {\em arXiv preprint}, volume arXiv:2407.12040. 2024.

\bibitem{ref10}
C.~T. Chien, R.~Y. Ju, K.~Y. Chou, and J.~S. Chiang.
\newblock Yolov9 for fracture detection in pediatric wrist trauma x‐ray images.
\newblock In {\em Electronics Letters}, volume~60, page e13248. 2024.

\bibitem{ref11}
M.~Yaseen.
\newblock What is yolov8: An in-depth exploration of the internal features of the next-generation object detector.
\newblock In {\em arXiv preprint}, volume arXiv:2408.15857. 2024.

\bibitem{ref12}
AI~Trends.
\newblock Yolov9 object detection with programmable gradient information (pgi) and generalized efficient layer aggregation network (gelan).
\newblock {\em Medium}, 2024.

\bibitem{ref13}
Roboflow.
\newblock Yolov9 deep dive.
\newblock {\em Roboflow Blog}, 2024.

\bibitem{ref14}
Chien-Yao Wang, Hong-Yuan~Mark Liao, Yueh-Hua Wu, Ping-Yang Chen, Jun-Wei Hsieh, and I-Hau Yeh.
\newblock Cspnet: A new backbone that can enhance learning capability of cnn.
\newblock In {\em Proceedings of the IEEE/CVF Conference on Computer Vision and Pattern Recognition Workshops (CVPRW)}, pages 390--391. 2020.

\bibitem{ref15}
Chien-Yao Wang, Hong-Yuan~Mark Liao, and I-Hau Yeh.
\newblock Designing network design strategies through gradient path analysis.
\newblock {\em Journal of Information Science and Engineering (JISE)}, 39(4):975--995, 2023.

\bibitem{ref16}
{Viso AI}.
\newblock Yolov9: Everything you need to know.
\newblock {\em Viso AI}, 2024.

\bibitem{ref17}
{XIS AI}.
\newblock Yolov9 vs yolov8: A comparative analysis.
\newblock {\em XIS AI Blog}, 2024.

\bibitem{ref18}
B.~Sener.
\newblock Yolov9 custom object detection.
\newblock {\em Medium}, 2024.

\bibitem{ref19}
{Ultralytics}.
\newblock Yolov9 model documentation.
\newblock {\em Ultralytics Docs}, 2024.

\bibitem{ref20}
Roboflow.
\newblock How to train a yolov9 model.
\newblock {\em Roboflow Blog}, 2024.

\bibitem{ref21}
{Datature}.
\newblock Yolov9: A comprehensive guide and custom dataset fine-tuning.
\newblock {\em Datature Blog}, 2024.

\bibitem{ref22}
Roboflow.
\newblock Yolov9 pytorch txt format description.
\newblock {\em Roboflow}, 2024.

\bibitem{ref23}
Muhammad Hussain.
\newblock Yolo-v5 variant selection algorithm coupled with representative augmentations for modelling production-based variance in automated lightweight pallet racking inspection.
\newblock {\em Big Data and Cognitive Computing}, 7(2):120, 2023.

\bibitem{ref24}
Rahima Khanam, Muhammad Hussain, Richard Hill, and Paul Allen.
\newblock A comprehensive review of convolutional neural networks for defect detection in industrial applications.
\newblock {\em IEEE Access}, 2024.

\bibitem{ref25}
M.~Hussain, A.~Razzaq, and H.~S.~Arif et~al.
\newblock Real-time patient indoor health monitoring and location tracking with optical camera communications on the internet of medical things, 2024.
\newblock ResearchGate, Aug. 2024.

\bibitem{ref26}
R.~Khanam and M.~Hussain.
\newblock What is yolov5: A deep look into the internal features of the popular object detector.
\newblock {\em arXiv preprint arXiv:2407.20892}, 2024.

\bibitem{ref27}
Tahir Hussain, Muhammad Hussain, Hussain Al-Aqrabi, Tariq Alsboui, and Richard Hill.
\newblock A review on defect detection of electroluminescence-based photovoltaic cell surface images using computer vision.
\newblock {\em Energies}, 16(10):4012, 2023.

\bibitem{aydin2023domain}
Burcu~Ataer Aydin, Muhammad Hussain, Richard Hill, and Hussain Al-Aqrabi.
\newblock Domain modelling for a lightweight convolutional network focused on automated exudate detection in retinal fundus images.
\newblock In {\em 2023 9th International Conference on Information Technology Trends (ITT)}, pages 145--150. IEEE, 2023.

\bibitem{hussain2023child}
Muhammad Hussain and Hussain Al-Aqrabi.
\newblock Child emotion recognition via custom lightweight cnn architecture.
\newblock In {\em Kids Cybersecurity Using Computational Intelligence Techniques}, pages 165--174. Springer, 2023.

\end{thebibliography}

\end{document}